# Contextual object categorization with energy-based model

Changyong Ri [1], Duho Pak [2], Cholryong Choe [3], Suhyang Kim [4], Yonghak Sin [5]

[1, 2, 3, 4, 5] *Institute of Information Science, Kim Il Song University, Pyongyang , D.P.R of Korea*

[5]sinyonghak@yahoo.com

**Abstract :** Object categorization is a hot issue of an image mining. Contextual information between objects is one of the important semantic knowledge of an image. However, the previous researches for an object categorization have not made full use of the contextual information, especially the spatial relations between objects. In addition, the object categorization methods, which generally use the probabilistic graphical models to implement the incorporation of contextual information with appearance of objects, are almost inevitable to evaluate the intractable partition function for normalization. In this work, we introduced fully-connected fuzzy spatial relations including directional, distance and topological relations between object regions, so the spatial relational information could be fully utilized. Then, the spatial relations were considered as well as co-occurrence and appearance of objects by using energy-based model, where the energy function was defined as the region-object association potential and the configuration potential of objects. Minimizing the energy function of whole image arrangement, we obtained the optimal label set about the image regions and addressed the evaluation of intractable partition function in conditional random fields. Experimental results show the validity and reliability of this proposed method.

**Keywords :** object categorization; contextual relations; energy-based model; conditional random field

# 1 Introduction

Object categorization which classifies image regions into semantic concepts generally is performed based on appearance (i.e. visual features) of the regions. As is well known, there is a semantic gap between low-level visual features and high-level semantics, so object categorization is a challenging task. However, human being correctly recognizes and classifies them through domain knowledge about certain scenes, which includes contextual information as well as appearance of objects. Contextual information about objects in images is an important knowledge to help the object recognition and classification [1, 9]. For example, when an object is isolated such as shown in (a) of Figure 1, the appearance of the object is not enough to recognize it as a specific category. In (b) of Figure1 where the whole scene is presented, we can recognize it as a boat with the contextual information about the object in the scene.

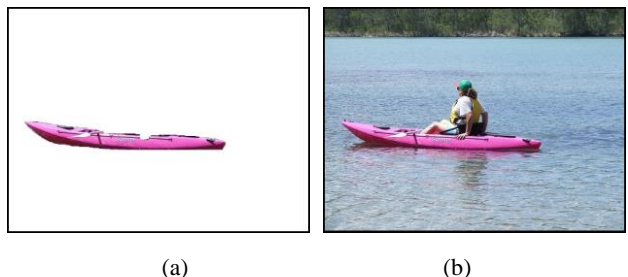

(a)        (b)

Figure1. (a) An object is isolated. The appearance of the object is not enough to classify it into specific category. (b) The whole scene. Based on contextual information about the object in the scene, we can recognize it as a boat.

Recently, many researchers have attempted to improve the accuracy of image categorization by incorporating contextual information with object's appearance [2, 3, 4, 5, 6, 7, 8]. They have incorporated co-occurrence and spatial arrangement with appearance of local objects. However, most of researchers have not made full use of the spatial arrangement of objects in an image. Therefore, it is challenging and significant for object categorization how to efficiently consider the spatial context with co-occurrence and appearance of objects. This is our first motivation for the approach to improve the contextual object categorization proposed in the rest of this presentation. In addition, the probabilistic graphical models (e.g. Markov Random Fields (MRFs) and Conditional Random Fields (CRFs)) have been employed to implement the incorporation of contextual information with appearance of objects in many previous works [3, 10, 11, 12, 13, 14]. However, there is an evaluation of intractable partition function for normalization of probabilistic models. Energy-based models (EBMs) capture dependencies between variables by associating a scalar energy to each configuration of the variables [15, 16, 18]. It allows us to relax the strict probabilistic modeling and to address the evaluation problem of intractable partition function. Therefore, how to appropriately capture energy functions of each configuration of the variables is our second motivation.

Concretely, we propose a two-phase strategy of object categorization, which includes the appearance-based initial labeling and context-based label refinement. Owing to the incorporating of contextual information with appearance of objects, the categorization accuracy could be improved. In our categorization model, the spatial relations, which are defined as fully-connected fuzzy spatial relations including directional, distance and topological relations between object regions, are considered as well as the co-occurrences by employing EBM, where we define an energy function with the region-object association potential and configuration potential of objects. Minimizing the energy function of whole image arrangement, we obtain the optimal label set about the image regions without the evaluation of intractable partition function in the probabilistic graphical models. Compared with the existing contextual object categorization models such as the literatures [3, 16], where the co-occurrence have merely used, and the literatures [12, 14], where have only used the fixed spatial relations and considered the simple frequency counting, we investigate the spatial arrangements in an image more sufficiently by defining fuzzy contextual relations. Moreover, the proposed EBM-based object categorization model overcomes the defects of the related works which rely on the use of undirected graphical models and often lead to intractable partition functions [3, 12]. The EBMs for object categorization were also used in [16, 18] similar to this work. However, their models did not fully reflect the spatial arrangement of an image. The contributions of our paper are as follows:

1) We propose fuzzy spatial relations for object categorization and so overcome the defects of related works which have only used the predefined and fixed spatial relations and considered the simple frequency counting as spatial constraints between objects. This can effectively reflect the contextual arrangement of an image and improve the categorization accuracy of objects.

2) We propose an EBM-based model for object categorization, which incorporates appearance with co-occurrence and spatial constraints of objects in an image. This overcomes the defects of the related works which rely on the use of undirected graphical models and often lead to intractable partition functions.

1.1 Related work

As mentioned above, traditional researches of object categorization have been mainly carried out using appearance features of local image regions. Appearance features such as color, texture, edge and shape cues can discriminate diversity of object categories in a certain extent. With the in-depth study, researchers have paid attention to the important role of contextual information for object categorization. A detailed survey of various contextual models for object categorization has been presented in the literatures [17, 18, 22].

There are two kinds of context level, namely global context and local context. Global context, which takes into account the object-scene interactions, can be used to restrict the possible objects that may be appeared in the scene [11, 24], while local context takes into account the interactions between objects [12], patches [2], or pixels [13]. Compared with global context, local context is easily accessible from training data, without expensive computations [17]. Moreover, when many different objects are appeared in an image, the contextual interactions between objects are most beneficial to capture information about the objects. Therefore, in this work we deal with an object-level contextual method.

There are three context types, namely semantic context (co-occurrence of objects), spatial context (spatial relations between objects) and scale context (size constraint for objects) [17]. A simple one of the most widely used contextual methods is to model the co-occurrence frequency of objects. Rabinovich et al. [3] and Felzenszwalb et al. [25] have taken into account information about how a target object co-occurs frequently with other objects and adjusted initial labels of segments assigned by using local detectors. The other one of contextual methods is to model the spatial relations between objects as well as the co-occurrence information. Object-level spatial contextual techniques capture information about the spatial configuration of objects as one of the sources that could infer the object categories. There are many approaches to exploit the spatial constraints between objects [10, 12, 14]. Zhou et al. [8] have extended the contextual methods to encode spatial relations between objects, where the spatial relations have been quantized to four predefined relations, namely above, below, inside and around. However, in most of the spatial contextual methods, the researchers have only used the predefined and fixed spatial relations and considered the simple frequency counting as spatial constraints between pair-wise objects. Therefore, in this work we deal with the fully-connected fuzzy spatial relations as well as the co-occurrence relations between objects in an image.

Generally, there is no causal relationship between image regions, so the undirected graphical models such as MRFs or CRFs are more suitable for modeling the interactions between object regions. Carbonetto et al. [2] considered information about the relations between image patches and proposed a MRF model that combined appearance feature vectors with spatial relations for object recognition. Heesch et al. [14] modeled the spatial and topological relations between objects in an image by employing the MRFs with asymmetric Markov parameters. To exploit both local features as well as contextual information, Torralba et al. [20] introduced Boosted Random Fields (BRFs) which used Boosting to learn the graph structure and local evidence of a CRF. Yuan et al. [10] employed simple grid-structure graphical models to describe the spatial dependencies between objects in an image. CRF model provides an approach to incorporate appearance and contextual information, and has the ability to directly predict the categories of regions, through modeling the conditional distribution and its relevance to inferring categories. Therefore, to model the contextual interactions between objects, in this work, we also introduce a fully-connected CRF model, where the nodes correspond to the region and the edges correspond to spatial relations between the object regions, and employ the EBM to handle the partition function of CRFs.

In other aspect, the post-processing strategy has been mainly adopted to improve the accuracy of object categorization, where the outputs of the object classifiers (i.e. initial labels) based on appearance features of regions are refined by applying different contextual models. Saathoff et al. [4] proposed a FCSP (fuzzy constraint satisfaction problems) for exploiting spatial prototypes and refined the initial labeling results from appearance-based classification with SVM. Papadopoulos et al. [18] introduced a genetic algorithm for refining of region labels using the confidence degrees from SVM classifiers as well as the spatial constraints between objects. Galleguillos et al. [12] firstly classified the image regions using a bag-of-features classifier, where each region was been assigned several candidate labels, then picked up a single label for each region using a CRF model, which incorporated confidence degrees of regions to object concepts with co-occurrence and spatial relations between objects. In this work, we also adopt two-phase strategy, namely appearance-based initial labeling and context-based label refinement, to improve the accuracy of object categorization.

Finally, we would like to specify the recently other approaches for exploring contextual information. For example, Lee et al. [21] introduced an object-graph descriptor for discovery of unknown object categories, which encode the object-level co-occurrence patterns; Singaraju et al. [27] developed a random field model for joint categorization and segmentation of objects, where they introduced the higher order potentials that encode the classification cost of a histogram extracted from objects belonged to different categories; Jain et al. [28] proposed a latent CRF model which captures the relations between features and visual words, relations between visual words and object categories, and spatial relations between visual words; Angin et al. [29] proposed a simple iterative algorithm for object categorization by exploiting the global co-occurrence frequencies of objects; Sun T et al. [30] proposed a object categorization method by combining local feature context with SVM classifiers.

Compared with the above mentioned researches, the method proposed in this work is easier to implement, more comprehensive to describe and better to perform the object categorization. Concretely, in order to improve the object categorization accuracy, we adopt a two-phase strategy and use the fuzzy SVM classifiers [23] for appearance-based initial labeling. Then, we propose the fuzzy spatial relations between objects to overcome the defects of related works which have only used the predefined and fixed spatial relations and considered the simple frequency counting as spatial constraints between objects. For object categorization, we employ CRF model which incorporates appearance with co-occurrence and spatial constraints of objects in an image. In addition, we introduce EBM, in which we define the region-object association potential and the configuration potential of objects, to overcome the defects of the related works which rely on the use of undirected graphical models and often lead to intractable partition functions. Experiments on four benchmark image datasets (LabelMe, SCEF, MSRC v2 and PASCAL VOC2010) show that the proposed method can improve the performance of object categorization compared with the state-of-the-art results.

## 2. Methodology

### 2.1. Problem formulation

Figure 2 depicts the overall flowchart of the contextual object categorization proposed in this work. The overall object categorization framework consists of two phases. The first one is the appearance-based region labeling. Firstly, the test images are manually or automatically segmented into object regions. In this work, we adopt manual and automatic segmentations together. While the image segmentation is no major topic of this work, we employ the advanced stability-based clustering method for automatic segmentation which proposed in Rabinovich et al. [3]. Then, the visual features of each segmented region are extracted. Afterward, the SVM classifiers are employed in order to classify image regions into semantic categories. The results of this classification are the initial labels and the confidence degrees of regions to object concepts (cf. Section 2.2).

The second one is the refinement of initial labels by using the contextual model. In this work, we employ the undirected graphical structure (i.e. CRF) for incorporating the appearance with co-occurrence and spatial relations between objects, and infer the final labels of each region by using energy-based model, which has ability to address the estimation of intractable partition functions in CRFs (cf. Section 2.4). Here, we take into account the fuzzy spatial relations as form of spatial context matrix, which has made full use of the spatial arrangement of objects in an image (cf. Section 2.3).

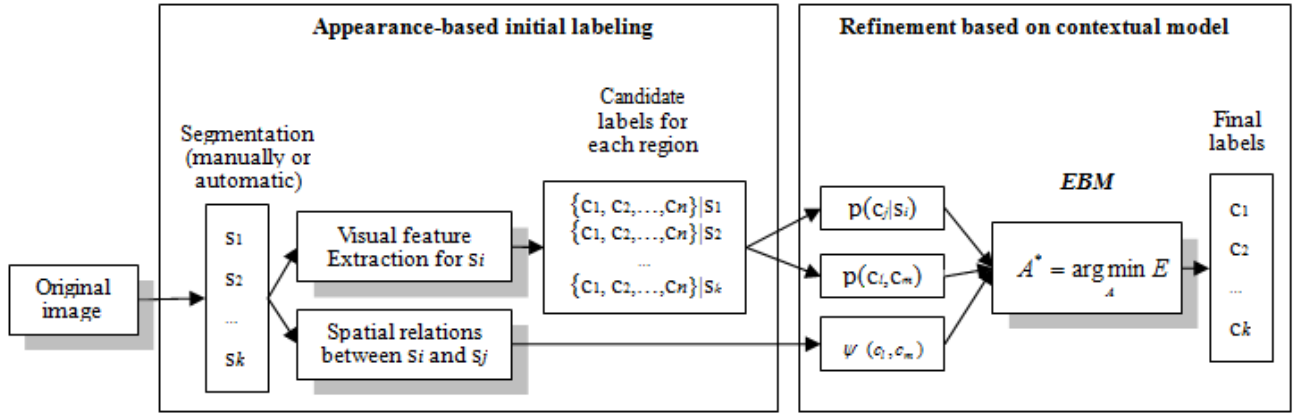

Figure2. The overall flowchart of contextual object categorization

2.2. Appearance-based initial labeling

Support Vector Machine (SVM) is one of the widely used techniques in semantic image processing owing to its powerful ability for classification and suitability for dealing with high-dimensional data. For initial region labeling, i.e. the assignment of object concepts to segmented regions based solely on appearance features, we employ SVM classifiers. The visual feature vectors of each segmented region $s_i$ are accepted as input data of SVMs, then we could obtain the concepts (i.e. candidate labels) $c_l$. Here, we use a non-linear Gaussian RBF kernel based on its performance in other pattern recognition applications [26]. Concrete process is as follows.

The conventional SVM had been originally proposed for the binary classification problem. Therefore, we adopt $n$ one-to-rest SVM classifiers for the purpose of the multiclass classification. Here, for the $l$th binary SVM, the separate plane classify the data $\mathbf{v}$ into class $c_l$ or non-class $c_l$. Let the decision function of the $l$th SVM (i.e. the SVM for the class $c_l$) is $D_l(\mathbf{v}) = \mathbf{w}_l \mathbf{v} + b_l$. The hyper plane $D_l(\mathbf{v}) = 0$ denotes the optimal separate plane. The data satisfied the equation $D_l(\mathbf{v}) = 1$ become the support vectors of class $c_l$, and the data satisfied the equation $D_l(\mathbf{v}) = -1$ become the support vectors of non-class $c_l$ (i.e. remaining class). If equation $D_l(\mathbf{v}) > 0$ is satisfied for one $l$th SVM, then $\mathbf{v}$ is clearly classified into the class $c_l$. However, if equation $D_l(\mathbf{v}) > 0$ is satisfied for plural $c_l$, or there is no $c_l$, then the classification of $\mathbf{v}$ is difficult.

To handle the unclassifiable data in the multiclass SVM, the fuzzy membership degree which denote the belonging degree of data $\mathbf{v}$ to class $c_l$, $\mu_{c_l}(\mathbf{v}) \in [0,1]$, is introduced. When the decision function for the class $c_l$ is equation (1), if $D_l(\mathbf{v}) \geq 1$, then the data $\mathbf{v}$ may be classified into the class $c_l$, and in this case the membership degree of the data $\mathbf{v}$ for the class $c_l$ is $\mu_{c_l}(\mathbf{v}) = 1$. If $D_l(\mathbf{v}) \leq -1$, the data $\mathbf{v}$ may be not classified into the class $c_l$, and $\mu_{c_l}(\mathbf{v}) = 0$. When $-1 \leq D_l(\mathbf{v}) \leq 1$, the data $\mathbf{v}$ may be partially classified into the class $c_l$, and the fuzzy membership degree is $\mu_{c_l}(\mathbf{v}) = (1 + D_l(\mathbf{v}))/2$. So we could perceive that the data lied in the hyper-plane of the $l$th SVM have the value of membership degree as 0.5. The above mentions are integrated as follows:

$$\mu_{c_l}(\mathbf{v}) = \begin{cases} 1, & \text{for } D_l(\mathbf{v}) \geq 1 \\ (1+D_l(\mathbf{v}))/2, & \text{for } -1 < D_l(\mathbf{v}) < 1 \\ 0, & \text{for } D_l(\mathbf{v}) \leq -1 \end{cases} \quad (1)$$

The unclassifiable regions include two cases: the first is that the data may be classified into plural classes, i.e. the membership degrees of the data **v** are 1 for several classes; the second is that the data may be not classified into any classes, i.e. the membership degrees of the data **v** are all 0. To resolve unclassifiable regions, we modify the membership degree as follows:

$$\hat{\mu}_{c_l}(\mathbf{v}) = \begin{cases} \dfrac{\mu_{c_l}(\mathbf{v})}{\sum_{r=1}^{n} \mu_{c_r}(\mathbf{v})}, & \text{for } \sum_{r=1}^{n} \mu_{c_r}(\mathbf{v}) \neq 0 \\ \dfrac{|D_l(\mathbf{v})|^{-1}}{\sum_{r=1}^{n} |D_l(\mathbf{v})|^{-1}}, & \text{for } \sum_{r=1}^{n} \mu_{c_r}(\mathbf{v}) = 0 \end{cases} \quad (2)$$

Later, we write the modified membership degree $\hat{\mu}_{c_l}(\mathbf{v})$ as $\mu_{c_l}(\mathbf{v})$ briefly, which means the belonging degree of data **v** to class $c_l$.

Using the fuzzy labeling method, we could determine the membership degrees of each class by equations (1) and (2). Thereby, the initial labels $Q_i = \{c_1, c_2, ..., c_n\}_i$ of each region $s_i \in I$ and their corresponding belief degrees $M_i = <\mu_{c_1}(s_i), \mu_{c_2}(s_i), ..., \mu_{c_n}(s_i)>$ are fuzzily obtained.

2.3. Fuzzy spatial relations between objects

The spatial relations between two objects can be divided into three classes: directional relations $R_1$, distance relations $R_2$ and topological relations $R_3$. The directional relations include *above*, *below*, *left* and *right*; the distance relations include *near* and *far*; the topological relations include *disjointed*, *bordering*, *invaded by* and *surrounded by* [31]. Moreover, these spatial relations can be combined into several classes, because the spatial relation between two object regions can be described by overlapping multiple relations, e.g. *invaded by* from *left*, *right* and *near*, etc.

Considering the characteristics of the natural scene images, i.e. *left* and *right* don't affect the object categorization, the directional relations can be divided into *above*, *below* and *beside*. Moreover, *near* and *far* are inverse each other, namely, the higher value of degree of *near*, the lower value of *far*. Therefore, the distance relations can be described by only *near* (or *far*) enough. Similarly, the topological relations can be described by only *surrounded by*. It means *disjointed* and *bordering* that the value of degree of *surrounded by* is 0. If the value of degree of *surrounded by* is greater than 0, then the topological relation becomes *invaded by* or *surrounded by*. Especially, if the value of degree of *surrounded by* is 1, then the topological relation is the complete *surrounded by*. Figure 3 shows the spatial relations between objects in an image.

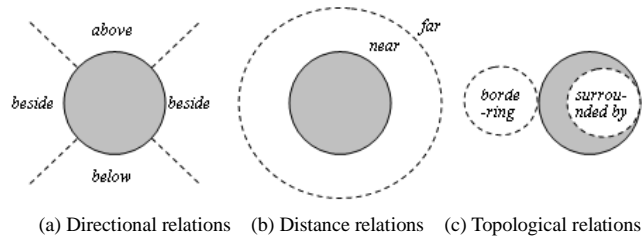

(a) Directional relations  (b) Distance relations  (c) Topological relations

Figure3. Spatial relations between objects in an image

A feature vector $\mathbf{r}_{ij}$ of spatial relation $R_{ij}$ between object regions $s_i$ and $s_j$ can be defined by the spatial relation descriptors $(\theta_{ij}, d_{ij}, \rho_{ij})$:

$$\mathbf{r}_{ij} = (r_{1,ij}, r_{2,ij}, r_{3,ij}) = (\mu_1(\theta_{ij}), \mu_2(d_{ij}), \mu_3(\rho_{ij})) \tag{3}$$

where, $\theta_{ij}$ denotes a angle between the horizontal axis and the line joining the centers of two object regions; $d_{ij}$ denotes minimum distance between the boundary pixels of two object regions; $\rho_{ij}$ denotes a ratio of the common perimeter between two object regions to the perimeter of the first object region.

In fact, the spatial relation between objects is a fuzzy relation. In equation (3), $\mu_R$ ($R = 1, 2, 3$) are the fuzzy membership degrees which denote the belonging degrees of the spatial relations between objects to the directional relations, distance relations and topological relations, respectively. The membership degrees of the five fuzzy spatial relations between objects can be computed with spatial relation descriptors as follows:

$$\mu_{ABOVE}(\theta_{ij}) = \begin{cases} \sin^2 \theta_{ij}, & \text{if } 0 < \theta_{ij} < \pi \\ 0, & \text{otherwise} \end{cases} \tag{4}$$

$$\mu_{BELOW}(\theta_{ij}) = \begin{cases} \sin^2 \theta_{ij}, & \text{if } -\pi < \theta_{ij} < 0 \\ 0, & \text{otherwise} \end{cases} \tag{5}$$

$$\mu_{BESIDE}(\theta_{ij}) = \cos^2 \theta_{ij} \tag{6}$$

$$\mu_{NEAR}(d_{ij}) = \frac{1}{1 + e^{\alpha_1(d_{ij} - \beta_1)}} \tag{7}$$

$$\mu_{SUR}(\rho_{ij}) = \frac{1}{1 + e^{-\alpha_2(\rho_{ij} - \beta_2)}} \tag{8}$$

where, $\alpha_1$ and $\alpha_2$ are the parameters that determine the crispness of the fuzzy membership degrees for distance relations and topological relations, respectively; $\beta_1$ is the cut-off value that divides the distance relations into *near* and *far* fuzzy relations; $\beta_2$ is the cut-off value that determines the *surrounded by* fuzzy relations.

Finally, the concrete directional relations between objects can be determined by maximum membership principle:

$$R_{1,ij} = \underset{W \in \{ABOVE, BELOW, BESIDE\}}{\arg\max} \mu_W(\theta_{ij}) \tag{9}$$

2.4. Region labeling with energy-based model

The EBMs capture the dependencies between variables by associating a scalar energy to each configuration of the variables. Here the correct configurations of variables generate minimum energy, while the incorrect configurations reveal higher values of energy. Therefore, the goal of the training phase is to obtain the energy function which associates low values of energy to correct configuration of the variables, and higher values of energy to incorrect one. In the inference phase, it is discovered that the values of the unobserved variables which minimize the energy. EBMs provide a unified framework for probabilistic graphical approaches (e.g. MRFs and CRFs). In contrast with MRFs and CRFs, the EBMs do not require the proper normalization and address the evaluation problem of intractable partition functions associated with estimating the normalization factor. Moreover, the absence of evaluation of the partition functions provides more flexibility in the modeling.

In this work, the region labeling is transformed into an energy minimizing problem by EBMs, where the energy function is obtained from integration of appearance, co-occurrence and spatial arrangement information of objects. An image $I$ which includes $k$ regions could be described as an undirected graph, where the nodes correspond to image regions and the edges correspond to dependencies between regions. Here all of the possible connections between nodes are considered. Through the region classification by appearance features, each region $s_i \in I$ in an image is associated with $n$ candidate labels, $Q_i = \{c_1, c_2, ..., c_n\}_i$, with the corresponding belief degrees, $M_i = <\mu_{c_1}(s_i), \mu_{c_2}(s_i), ..., \mu_{c_n}(s_i)>$. The association of region $s_i$ with label $c_l$ is represented by $a(s_i) = c_l$ (or $a_i = c_l$ simply), where $a$ is an association variable. Then the region labeling of whole image is expressed as $A = \{a_1, a_2, ..., a_k\}$, where each region is associated with one of its candidate labels, i.e. the values of the association variables. In form, the variable $a_i$ is the random variable of corresponding region $s_i$, and it's value is as $c_l \in Q_i$. An example of this graphical model is shown in Figure 4, where $k = 5$. Now, using the EBMs, the goal of the region labeling of image $I$ is to find the best value set $A^*$ of association variables, which minimizes the energy.

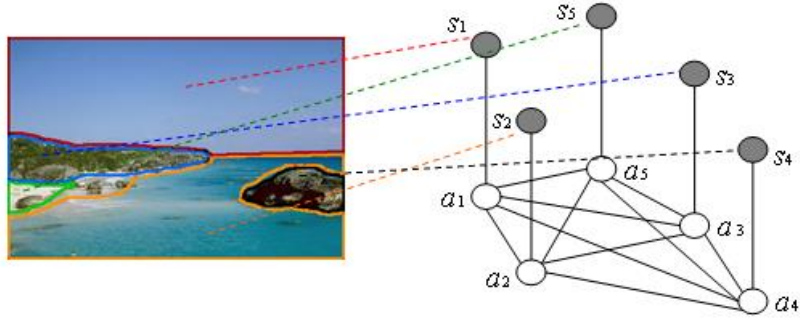

Figure4. Energy-based model with fully-connected CRF. The region-object association potential is defined on the vertical lines, and the configuration potential is defined on the horizontal lines.

In order to find the best value set $A^*$, i.e. the best configuration of labels for image $I$, we define the energy function which measures the suitability of any configuration of labels:

$$E(A) = -\left( \lambda_1 \cdot \sum_{i=1}^{k} \Phi(a_i) + \sum_{\substack{i,j=1 \\ (i \neq j)}}^{k} \Psi(a_i, a_j) \right) \quad (10)$$

Where, $\Phi(a_i)$ denotes a region-object association potential, which is depended on the label $a(s_i)$ of region $s_i$ classified with appearance features of the region; $\Psi(a_i, a_j)$ denotes a configuration potential of objects, which reflects to a interaction between labels $a(s_i)$ and $a(s_j)$ assigned to related regions $s_i$ and $s_j$; $\lambda_1$ denotes a factor that adjusts the contribution of the region-object association potential compared with the configuration potential of objects.

**Region-object association potential** $\Phi(a_i)$ is described by taking into account the posterior probability $p(c_l | s_i)$ exported from the appearance-based object classifiers as well as the prior probability $p(c_l)$ of object occurrence:

$$\Phi(a_i) = \lambda_2 \cdot p(c_l | s_i) + p(c_l) \quad (11)$$

Where, the parameter $\lambda_2$ is a factor that adjusts the influence of the appearance-based classifiers compared with the concept occurrence. Using discriminative classifiers such as SVMs with appearance features of local regions, the every candidate label $c_l$ for each region $s_i$ is afforded the corresponding belief degree $\mu_{c_l}(s_i)$, which denote the belonging degree of the region to its class. Therefore, it is natural that the posterior probability by appearance-based object classifiers is as follows:

$$p(c_l | s_i) = \mu_{c_l}(s_i) \quad (12)$$

The prior probability of object occurrence could be obtained by frequency count of the object concept over all regions in training dataset.

**Configuration potential of objects** $\Psi(a_i, a_j)$ reflects to a interaction between labels $a(s_i)$ and $a(s_j)$ assigned to related regions $s_i$ and $s_j$. The labels considered in region labeling are the semantic concepts of objects, so the potential $\Psi(a_i, a_j)$ denotes the contextual interactions between objects, which include spatial context and semantic context (i.e. co-occurrence), and is expressed as follows:

$$\Psi(a_i, a_j) = \lambda_3 \cdot \psi_r(c_l, c_m) + p(c_l, c_m) \cdot p(c_l | s_i) \quad (13)$$

Where, the parameter $\lambda_3$ is a factor that adjusts the influence of spatial context compared with co-occurrence; the term $p(c_l | s_i)$ is added in order to weight the influence of the confidence of neighboring labels. The spatial context interaction $\psi(c_l, c_m)$ is obtained using the membership degrees of fuzzy spatial relations between objects. Considering that the fuzzy spatial relation between objects $c_l$ and $c_m$ is characterized by feature vector according to equation (3), where the members of vector denote the membership degrees $\mu_r(c_l, c_m)$, the spatial context interaction $\psi(c_l, c_m)$ is estimated by using a Euclidean distance-based formulation:

$$\psi(c_l, c_m) = 1 - |\bar{\mathbf{r}}_{lm} - \mathbf{r}_{ij}| \quad (14)$$

where, $\bar{\mathbf{r}}_{lm}$ is the mean vector of spatial relations between concept pair $c_l$ and $c_m$ over all images in training set, while $\mathbf{r}_{ij}$ is the feature vector of spatial relations between regions $s_i$ and $s_j$, candidate labels of which are $c_l$ and $c_m$ respectively. Additionally, the semantic context interaction $p(c_l, c_m)$ is estimated by frequency count of the co-occurrence of object pairs over all images in training set. The co-occurrence matrices for four datasets (i.e. LabelMe, SCEF, MSRC v2 and PASCAL VOC2010) which have been used in this work are shown in Figure 5.

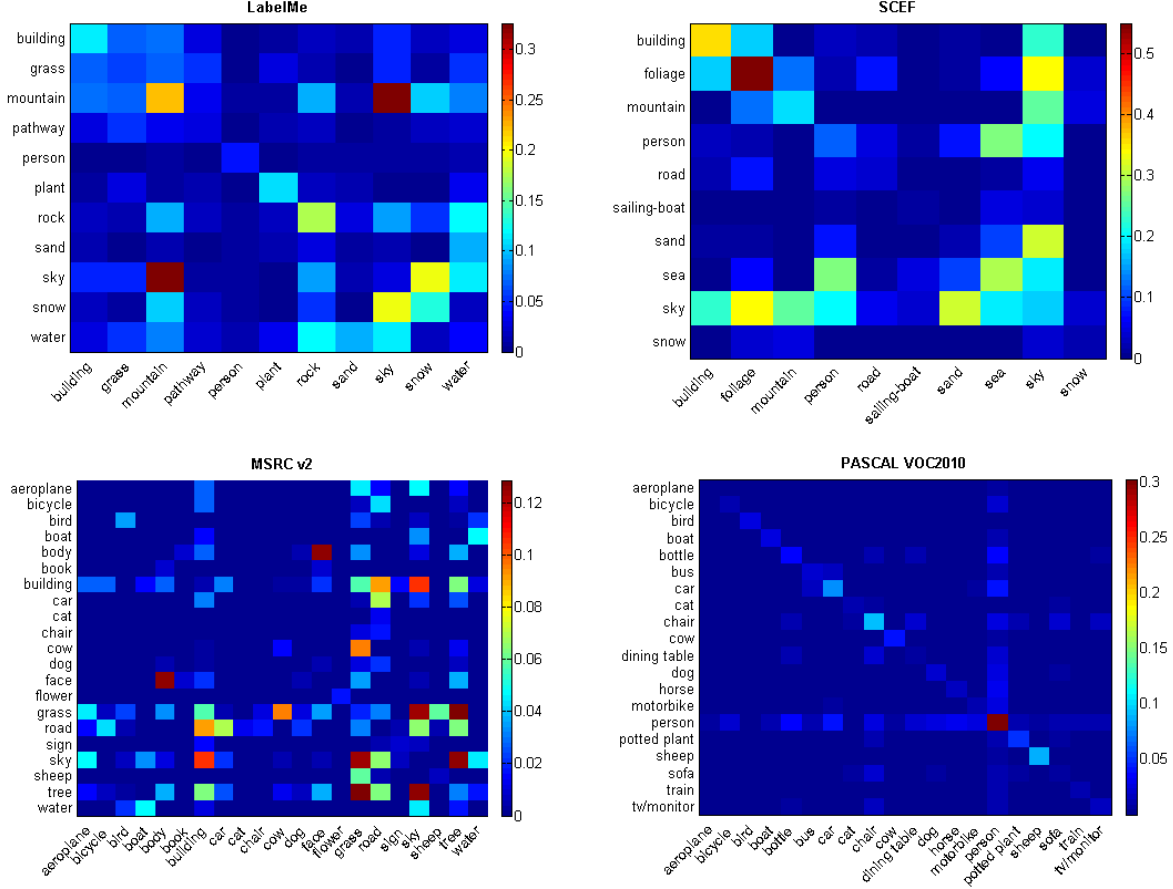

Figure5. Co-occurrence frequency between object pairs in four datasets

Therefore, the energy function could be rewritten as follows:

$$E(A) = -\left( \sum_{i=1}^{k} \alpha \cdot p(c_l | s_i) + \beta \cdot p(c_l) + \sum_{\substack{i,j=1 \\ (i \neq j)}}^{k} \delta \cdot \psi(c_l, c_m) + p(c_l, c_m) \cdot p(c_l | s_i) \right) \quad (15)$$

There are three adjustable parameters in the energy function, namely $\alpha$, $\beta$ and $\delta$. These parameters are selected by trial on a validation dataset.

The region labeling is simplified to find the set of concepts which minimize the energy such as equation (14). At the inference stage, we take as input of EBM the appearance-based classification results (i.e. posterior probabilities), as well as the fuzzy spatial relations between object regions. Then, ensuring the minimum of overall energy value, a particular object concept is assigned to every region. The Iterated Conditioned Modes (ICM) algorithm [19], which is a method commonly used in graphical models such as CRFs and MRFs, is generally employed to find the minimum energy configuration.

$$A^* = \arg\min_{A} E(A) \quad (16)$$

# 3. Experimental results

## 3.1. Experimental setup

In order to verify the effectiveness of object categorization method proposed in this work, we carried out the related experiments with two fields: manually annotated regions and automatically classified regions. We used four benchmark image datasets such as LabelMe[1], SCEF[2], MSRC v2[3] and PASCAL VOC2010[4]. We divided each dataset into two parts, i.e. training dataset and test dataset. Table 1 illustrates the number of total, training and test datasets, the number of segmented regions and the supported object concepts for four datasets.

Table1. The number of total, training and test datasets, the number of segmented regions and the supported object concepts for four datasets used in these experiments

| Dataset | Number of images | | | Number of Regions | Supported concepts |
|---|---|---|---|---|---|
| | total | training | test | | |
| LabelMe | 1093 | 400 | 694 | 6446 | 11 concepts: building, grass, mountain, pathway, person, plant, rock, sand, sky, snow, water |
| SCEF | 922 | 400 | 522 | 6244 | 10 concepts: building, foliage, mountain, person, road, sailing-boat, sand, sea, sky, snow |
| MSRC v2 | 591 | 225 | 256 | 2062 | 21 concepts: aeroplane, bird, boat, body, book, building, bicycle, car, cat, chair, cow, dog, face, flower, grass, road, sheep, sign, sky, tree, water |
| PASCAL VOC2010 | 1853 | 897 | 956 | 4508 | 20 concepts: aeroplane, bicycle, bird, boat, bottle, bus, car, cat, chair, cow, dining table, dog, horse, motorbike, person, potted plant, sheep, sofa, train, tv/monitor |

For each dataset, we used the training set to train the multiclass fuzzy SVM classifiers. Probabilities of co-occurrence and fuzzy spatial relations between object concepts were computed using the training set as well. As mentioned in Section 2.2, we used the SVM classifiers with a non-linear Gaussian RBF kernel to obtain the initial labels and corresponding belief degrees. The values for kernel radiuses $\sigma$ and cost parameteres [26] $c$ of SVM are shown in table 2, which have been slected empirically whith 10-fold cross validation to maximize the average performance of classification, and the other parameter values used in this experiment are also were shown in table 2.

Table2. The parameter values used in these experiments

| Parameter | $\alpha_1$ | $\beta_1$ | $\alpha_2$ | $\beta_2$ | $\alpha$ | $\beta$ | $\delta$ | $\sigma$ | $c$ |
|---|---|---|---|---|---|---|---|---|---|
| Value | 20 | 0.25 | 10 | 0.6 | 1.4 | 0.3 | 0.8 | 2 | 10 |

While the image segmentation is no major topic of this work, we employ the advanced stability-based clustering method for automatic segmentation which proposed in Rabinovich et al. [3]. Then, the visual features of each segmented region are extracted. For LabelMe dataset, we use the visual features which are the concatenations of a 54-bin linear HSV color histogram, an 8-bin edge direction histogram and the 24 features of the gray-level co-occurrence matrix: contrast, energy, entropy, homogeneity, inverse difference moment and correlation for the displacements $\overrightarrow{1,0}$, $\overrightarrow{1,1}$, $\overrightarrow{0,1}$ and $\overrightarrow{-1,1}$ [32]. For SCEF, MSRC v2 and PASCAL VOC2010 datasets, we use the MPEG-7 and the SIFT descriptors as visual features, which have been used in relative baselines.

The performance is evaluated with categorization accuracies which defined as the percentage of correctly classified regions. The reported results are the average values of them.

---

[1] http://labelme.csail.mit.edu/
[2] http://mklab.iti.gr/project/scef
[3] http://research.microsoft.com/en-us/projects/objectclassrecognition/
[4] http://pascallin.ecs.soton.ac.uk/challenges/VOC/voc2010/index.html

3.2. Results and discussion

In order to evaluate the importance of contextual information in an image and the effectiveness of EBM-based contextual model, we compared the results of proposed method with the results of non-contextual method and CRF-based method for LabelMe dataset. Figure 6 illustrates the comparision of object classification accuracies of these three methods. It can be seen that the EBM-based method outperform the non-contextual method and the CRF-based method. The average categorization accuracy is 50.09% for EBM-based method, while the one is 46.19% for non-contextual method and is 49.53% for CRF-based method. Specifically, compared with the non-contextual method, the categorization accuracy incresed in most of the 11 categories by using contextual information, i.e. an increase from 1%-12%, besides the category "snow" in which appeared small decrease. Compared with the CRF-based method, the average accuracy incresed 0.6%, because all of these two methods used the co-occurrence and spatial information between objects with appearance features.

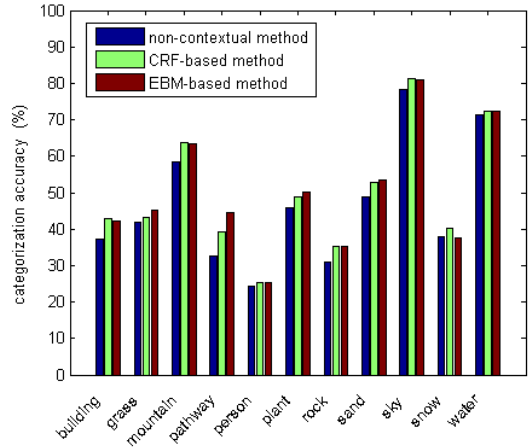

Figure6. Comparison of object categorization accuracies for LabelMe dataset

In order to verify the effectiveness of the proposed method, we compared the object categorization results with several state-of-the-art baselines, namely Galleguillos et al [12], Escalante et al [16] and Papadopoulos et al [18] for SCEF, MSRC v2 and PASCAL VOC2010 image datasets. Here, we adopted the MPEG-7 and the SIFT descriptors which have been used in the baselines as visual features extracted from each segmented region. Table 3 shows the comparison of average categorization accuracies with the baselines.

Table3. Comparison of the average categorization accuracy with several state-of-the-art baselines (%)

| Datasets<br>Methods | SCEF | | MSRC | | PASCAL | |
|---|---|---|---|---|---|---|
| | MPEG-7 | SIFT | MPEG-7 | SIFT | MPEG-7 | SIFT |
| CRF-based method by Galleguillos et al. [12] (2008) | | | | 68.38 | | 36.70 |
| EBM-based method by Escalante et al [16] (2011) | 60.35 | | | | | |
| EBM-based method by Papadopoulos et al. [18] (2011) | 58.58 | 65.57 | 43.24 | 44.31 | 21.07 | 30.57 |
| EBM-based method proposed in this work | **61.91** | **65.95** | **59.90** | **68.81** | **36.78** | **37.45** |

In the CRF-based method by Galleguillos et al [12], the contextual information, i.e. location and co-occurrence have been incorporated with appearance of objects by using CRF model. Here, they used only four fixed spatial relations such as *Above*, *Below*, *Inside* and *Around*, and approximated the partition function by using the Monte Carlo integration. In contrast, Escalante et al [16] and Papadopoulos et al [18] introduced the EBM-based models which avoid the evaluation of intractable partition function. In Escalante et al, the energy function was defined as integrating the appearance-based observation and the simple concept co-occurrence statistics. In Papadopoulos et al, the fuzzy spatial relations between objects have been used for defining energy function as weel as the co-occurrence and appearance of objects. However, they merely used eight directionoal relations between objects such as *Above*, *Right*, *Below*, *Left*, *Below-Right*, *Below-Left*, *Above-Right* and *Above-Left*, and did not fully reflect the spatial arrangement of an image. In proposed method, the using of fuzzy spatial relations and EBM-based categorization model overcome the defects of related works in incorporating spatial contextual information for object categorization. As shown in Table 3, the proposed method outperforms the state-of-the-art baselines.

Figure 7 illustrates the categorezation accuracies per object categories for SCEF, MSRC v2 and PASCAL

VOC2010 datasets. It can also be seen that several categories such as "road" and "sailling-boat" in SCEF dataset and "aeroplane" in MSRC v2 dataset exhibit low categorization accuracies, i.e. lower than 50%. It is due to the fact that the results of appearance-based classification of these categories are insufficient. For PASCAL dataset, the effect of the incorporation of contextual information between objects is not obvious, and for most categories the categorization accuracies are relatively low. It is due to the fact that the images of this dataset contain very few objects, and in this case the contextual relations between objects are not significant. Thus it can be seen that, the effectiveness of contextual information in improving the appearance-based initial labeling results is highlighted under presence of many objects in an image.

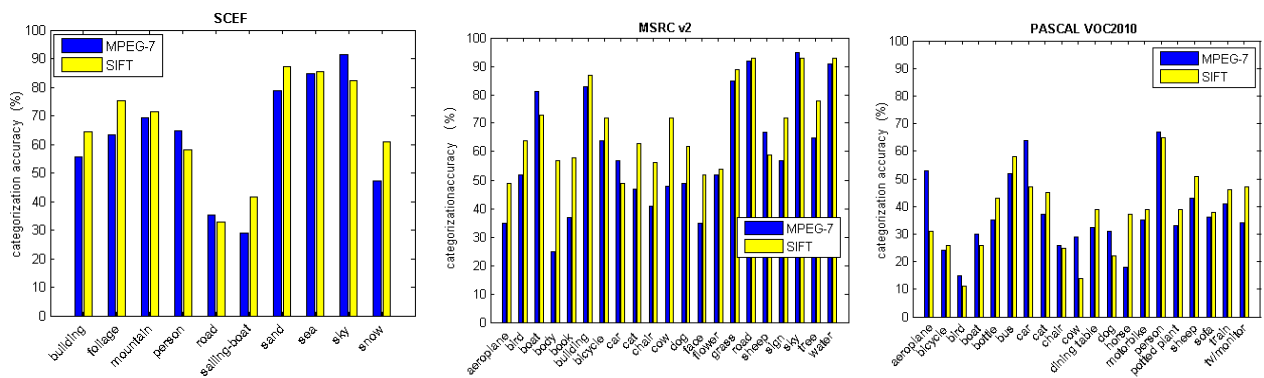

Figure7. Categorization accuracies per object categories for SCEF, MSRC v2 and PASCAL VOC2010 datasets

Finaly, we investigated the effect of the number of candidate labels on the final categorization accurecies. Figure 8 illustrates the curves of average accuracies that have been obtained by taking into account the top $n$ labels which sorted by its confidence degree. The experimental results show that, when the number of candidate labels less than 3, for all image datasets the accuracies were improved with the increasing of the number, and when the number of candidate labels greater than 5 (or 8 for MSRC dataset), the accuracies were converged. However, when the number of candidate labels is increased, the computational complexity is increased too. This illustrates that the suitable number of cadidate labels could lead to wishful result by using the proposed contextual model.

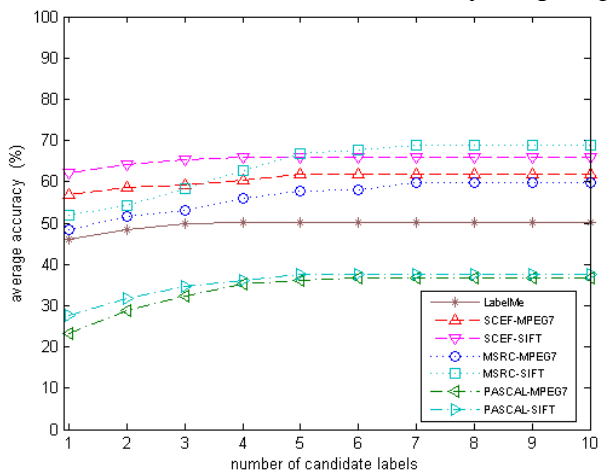

Figure 8. Change curve of the average categorization accuracy with the change of the number of candidate labels

Through the comparison analysis of experiments, we have verified the suitability and effectiveness of the proposed contextual object categorization method with energy-based model. The proposed method makes full use of the spatial relational information between objects in an image. In addition, it employs the EBM which incorporates appearance with co-occurrence and spatial constraints of objects. Therefore, it improves the performance of object categorization.

## 4. Conclusions

The full use of the contextual information as well as the appearence of objects is an important method to improve the accuracy of object categorization. In this paper, we proposed a contextual object categorization method, which used the fuzzy spatial relations between objects in an image. This effectively reflects the contextual arrangement of an image. In addition, the proposed method employed the EBM for object categorization, which incorporates appearance with co-occurrence and spatial constraints of objects, in order to address the evaluation of partition function in CRFs. Therefore we have improved the accuracy of object categorization. Experimental results showed the validity and the suitability of the proposed method.

However, the contextual information of an image inclides all possible knowledge, namely scale (size) context as well as co-occurrence and spatial constraints employed in this work. In further, we will research to combine the different contextual information such as the scale context with object categotization system so as to achieve better recognition results.

**Acknowledgements** This work was partly supported by the 973 Program (2013CB329504), NSF of China (No. 61272261), NSF of Zhejiang (Y1110152), and STD of Zhejiang (2012C21002).

## References

[1] Oliva A, Torralba A. The role of context in object recognition[J]. Trends in cognitive sciences, 2007, 11(12): 520-527.
[2] Carbonetto P, de Freitas N, Barnard K. A statistical model for general contextual object recognition[M]//Computer Vision-ECCV 2004. Springer Berlin Heidelberg, 2004: 350-362.
[3] Rabinovich A, Vedaldi A, Galleguillos C, et al. Objects in context[C]//Computer Vision, 2007. ICCV 2007. IEEE 11th International Conference on. IEEE, 2007: 1-8.
[4] Saathoff C, Staab S. Exploiting spatial context in image region labelling using fuzzy constraint reasoning[C]//Image Analysis for Multimedia Interactive Services, 2008. WIAMIS'08. Ninth International Workshop on. IEEE, 2008: 16-19.
[5] Yang L, Zheng N, Yang J. A unified context assessing model for object categorization[J]. Computer Vision and Image Understanding, 2011, 115(3): 310-322.
[6] Cinbis R G, Sclaroff S. Contextual object detection using set-based classification[M] //Computer Vision–ECCV 2012. Springer Berlin Heidelberg, 2012: 43-57.
[7] Choi M J, Torralba A, Willsky A S. A tree-based context model for object recognition[J]. Pattern Analysis and Machine Intelligence, IEEE Transactions on, 2012, 34(2): 240-252.
[8] Zhou C, Liu C. Semantic image segmentation using low-level features and contextual cues[J]. Computers & Electrical Engineering, 2013.
[9] Parikh D, Zitnick C L, Chen T. Exploring Tiny Images: The Roles of Appearance and Contextual Information for Machine and Human Object Recognition[J]. Pattern Analysis and Machine Intelligence, IEEE Transactions on, 2012, 34(10): 1978-1991.
[10] Yuan J, Li J, Zhang B. Exploiting spatial context constraints for automatic image region annotation[C]//Proceedings of the 15th international conference on Multimedia. ACM, 2007: 595-604.
[11] Verbeek J, Triggs W. Scene segmentation with crfs learned from partially labeled images[J]. 2007.
[12] Galleguillos C, Rabinovich A, Belongie S. Object categorization using co-occurrence, location and appearance[C]//Computer Vision and Pattern Recognition, 2008. CVPR 2008. IEEE Conference on. IEEE, 2008: 1-8.
[13] Shotton J, Winn J, Rother C, et al. Textonboost for image understanding: Multi-class object recognition and segmentation by jointly modeling texture, layout, and context[J]. International Journal of Computer Vision, 2009, 81(1): 2-23.
[14] Heesch D, Petrou M. Markov random fields with asymmetric interactions for modelling spatial context in structured scene labelling[J]. Journal of Signal Processing Systems, 2010, 61(1): 95-103.
[15] Y. LeCun, S. Chopra, R. Hadsell, M.A. Ranzato, F.J. Huang, Energy-based models, in: Predicting Structured Data, MIT Press, 2007, pp. 191–246 (Chapter 10).
[16] Escalante H J, Montes-y-Goméz M, Sucar L E. An energy-based model for region-labeling[J]. Computer vision and image understanding, 2011, 115(6): 787-803.
[17] Galleguillos C, Belongie S. Context based object categorization: A critical survey[J]. Computer Vision and Image Understanding, 2010, 114(6): 712-722.
[18] Papadopoulos G T, Saathoff C, Escalante H J, et al. A comparative study of object-level spatial context techniques for semantic image analysis[J]. Computer Vision and Image Understanding, 2011, 115(9): 1288-1307.
[19] Besag J. On the statistical analysis of dirty pictures[J]. Journal of the Royal Statistical Society. Series B (Methodological), 1986: 259-302.
[20] Torralba A, Murphy K P, Freeman W T. Contextual models for object detection using boosted random fields[C]//Advances in neural information processing systems. 2004: 1401-1408.
[21] Lee Y J, Grauman K. Object-graphs for context-aware category discovery[C]//Computer Vision and Pattern Recognition (CVPR), 2010 IEEE Conference on. IEEE, 2010: 1-8.
[22] Divvala S K, Hoiem D, Hays J H, et al. An empirical study of context in object detection[C]//Computer Vision and Pattern

Recognition, 2009. CVPR 2009. IEEE Conference on. IEEE, 2009: 1271-1278.
[23] Ri C Y, Yao M. Semantic Image Segmentation Based on Spatial Context Relations[C]//Information Science and Engineering (ISISE), 2012 International Symposium on. IEEE, 2012: 104-108.
[24] Russell B, Torralba A, Liu C, et al. Object recognition by scene alignment[C]//Advances in Neural Information Processing Systems. 2007: 1241-1248.
[25] Felzenszwalb P F, Girshick R B, McAllester D, et al. Object detection with discriminatively trained part-based models[J]. Pattern Analysis and Machine Intelligence, IEEE Transactions on, 2010, 32(9): 1627-1645.
[26] Serrano N, Savakis A E, Luo J. Improved scene classification using efficient low-level features and semantic cues[J]. Pattern Recognition, 2004, 37(9): 1773-1784.
[27] Singaraju D, Vidal R. Using global bag of features models in random fields for joint categorization and segmentation of objects[C]//Computer Vision and Pattern Recognition (CVPR), 2011 IEEE Conference on. IEEE, 2011: 2313-2319.
[28] Jain A, Zappella L, McClure P, et al. Visual dictionary learning for joint object categorization and segmentation[M]//Computer Vision–ECCV 2012. Springer Berlin Heidelberg, 2012: 718-731.
[29] Angin P, Bhargava B. A Confidence Ranked Co-Occurrence Approach for Accurate Object Recognition in Highly Complex Scenes[J]. Journal of Internet Technology, 2013, 14(1): 13-19.
[30] Sun T, Zhang C, Liu J, et al. Object Categorization Using Local Feature Context[M]//Advances in Multimedia Modeling. Springer Berlin Heidelberg, 2013: 327-333.
[31] Aksoy S, Tusk C, Koperski K, et al. Scene modeling and image mining with a visual grammar[J]. Frontiers of remote sensing information processing, 2003: 35-62.
[32] Cheng H, Wang R. Semantic modeling of natural scenes based on contextual Bayesian networks[J]. Pattern Recognition, 2010, 43(12): 4042-4054.